\documentclass[sn-standardnature]{sn-jnl}

\theoremstyle{thmstyleone}%
\theoremstyle{thmstyletwo}%

\theoremstyle{thmstylethree}%

\usepackage{array}
\usepackage{arydshln}
\usepackage{tabularray}

\usepackage{natbib}
\setcitestyle{numbers}
\setcitestyle{square}

\usepackage{booktabs}
\newcommand{\ra}[1]{\renewcommand{\arraystretch}{#1}}
\usepackage{multirow}

\usepackage{tabularx}
\usepackage{tabu}
\begin{document}


\title[MC-NN for IAV Prediction]{MC-NN: An End-to-End Multi-Channel Neural Network Approach for Predicting Influenza A Virus Hosts and Antigenic Types}


\author*[1]{\fnm{Yanhua} \sur{Xu}}\email{Y.Xu137@liverpool.ac.uk}

\author[1]{\fnm{Dominik} \sur{Wojtczak}}\email{D.Wojtczak@liverpool.ac.uk}

\affil*[1]{\orgdiv{Department of Computer Science}, \orgname{University of Liverpool}, \city{Liverpool}, \postcode{L69 3BX}, \state{Merseyside}, \country{UK}}


\abstract{Influenza poses a significant threat to public health, particularly among the elderly, young children, and people with underlying diseases. The manifestation of severe conditions, such as pneumonia, highlights the importance of preventing the spread of influenza. An accurate and cost-effective prediction of the host and antigenic subtypes of influenza A viruses is essential to addressing this issue, particularly in resource-constrained regions. In this study, we propose a multi-channel neural network model to predict the host and antigenic subtypes of influenza A viruses from hemagglutinin and neuraminidase protein sequences. Our model was trained on a comprehensive data set of complete protein sequences and evaluated on various test data sets of complete and incomplete sequences. The results demonstrate the potential and practicality of using multi-channel neural networks in predicting the host and antigenic subtypes of influenza A viruses from both full and partial protein sequences.}

\keywords{Influenza, CNN, BiGRU, Transformer, Deep Learning}



\maketitle


\section{Introduction}\label{sec1}
The impact of influenza viruses on respiratory diseases worldwide is substantial, leading to severe infections in the lower respiratory tract, hospitalisations, and mortality. There are estimated to be $>$ 5 million hospitalisations annually due to influenza-related respiratory illnesses \cite{lafond2021global}. The incidence of severe influenza-associated diseases and hospitalisation is highest among individuals at the extremes of age and those with pre-existing medical conditions. The virus spreads primarily through droplets, aerosols or direct contact, and up to 50\% of infections are asymptomatic \cite{lau2010viral, wilde1999effectiveness}. The influenza virus can cause various complications associated with high fatality rates, including secondary bacterial pneumonia, primary viral pneumonia, chronic kidney disease, acute renal failure, and heart failure \cite{watanabe2013renal, casas2018aggressive, public2020influenza}.

The influenza virus’s genome is comprised of single-stranded ribonucleic acid (RNA) segments. It is classified into four genera differentiated primarily by the antigenic properties of the nucleocapsid (NP) and matrix (M) proteins \cite{shaw2013orthomyxoviridae}. Currently, the influenza virus has four types: A (IAV), B (IBV), C (IVC), and D (IVD). Among them, IAV is the most widespread and virulent, capable of triggering major public health disruptions and pandemics, as demonstrated by the Spanish Flu of 1918–1919 that resulted in an estimated 20–100 million deaths \cite{mills2004transmissibility}. IAV is further subtyped by the antigenic properties of its hemagglutinin (HA) and neuraminidase (NA) surface glycoproteins, with 18 HA and 11 NA subtypes currently known \cite{asha2019emerging}. The avian influenza viruses, including H5N1, H5N2, H5N8, H7N7, and H9N2, can also spread from birds to humans with potentially deadly consequences, although this rarely occurs.

The HA and NA proteins of the influenza virus play a crucial role in its ability to infect host cells by allowing it to recognise and attach to specific receptors on host epithelial cells, followed by replication and release into neighbouring cells through the action of NA \cite{james2017influenza}. The immune system can respond to the virus by attacking and destroying infected tissue, although death can sometimes result from organ failure or secondary infections. The continuous evolution of the virus through point mutations in the genes encoding HA and NA can result in antigenic drift, leading to seasonal influenza, or the rarer antigenic shift, resulting in the emergence of new viruses with a significant change in HA and NA production that can trigger pandemics \cite{clayville2011influenza}.

In this study, we aim to predict IAV subtypes and hosts using a multi-channel neural network (MC-NN) approach comprising a combination of convolutional neural networks (CNNs), bidirectional gated recurrent units (BiGRUs), and transformer models.  The models are trained on a large-scale integrated protein sequence data set collected before 2020 and evaluated on both a post-2020 data set and a data set containing incomplete sequences. The study includes a broad range of hosts. Its results demonstrate the superiority of our multi-channel approach, with the transformer model achieving  83.39\%, 99.91\% and 99.87\% F\textsubscript{1} scores for the host, HA subtype and NA subtype prediction, respectively, in the post-2020 data set. Furthermore, its performance on incomplete sequences reached 76.13\%, 95.37\% and 96.37\% F\textsubscript{1} scores for the host, HA subtype and NA subtype prediction, respectively.

\section{Related Work}

The detection of IAV hosts and subtypes can enhance the surveillance of influenza and mitigate its spread. However, traditional methods for virus subtyping, such as nucleic acid-based tests (NATs), are labour-intensive and time-consuming \cite{vemula2016current}. To address this issue, researchers have explored various supervised machine learning-based methods for predicting IAV hosts or subtypes. These include using CNNs \cite{clayville2011influenza, fabijanska2019viral, scarafoni2019predicting}, support vector machines (SVM) \cite{ahsan2018first, xu2017predicting, kincaid2018n}, decision trees (DT) \cite{ahsan2018first, attaluri2009integrating}, and random forests (RF) \cite{kincaid2018n, eng2014predicting, kwon2020study}.

In order to train machine learning models, the protein sequences need to be transformed into numerical vectors. This transformation has been achieved through various methods, including one-hot encoding \cite{clayville2011influenza, eng2014predicting, mock2021vidhop}, pre-defined binary encoding schemes \cite{attaluri2010applying}, ASCII codes \cite{fabijanska2019viral}, Word2Vec \cite{xu2017predicting}, and the use of physicochemical features \cite{chrysostomou2021classification, sherif2017classification, kwon2020study, yin2020hopper}. However, using handcrafted feature sets or physicochemical features requires a feature selection process, which can be time-consuming. This study used word embedding to allow the models to learn features from the training data since this approach is more convenient and efficient. Previous studies have focused on either higher classification (i.e. avian, swine, or human) or a single class of hosts from a single database. In contrast, this study collects data from multiple databases and focuses on a broad range of hosts.

MC-NNs have been used in various applications, such as face detection \cite{george2019biometric}, relation extraction \cite{chen2020multi}, entity alignment \cite{cao2019multi}, emotion recognition \cite{yang2018emotion}, and haptic material classification \cite{kerzel2017haptic}. To our knowledge, few studies have used MC-NNs for infectious disease predictions. In this study, we propose using three MC-NN architectures to simultaneously predict IAV hosts and subtypes rather than training separate models for each task.

\section{Materials and Methods}
\subsection{Data Preparation}
\subsubsection{Hemagglutinin and Neuraminidase Protein Sequences}
The hemagglutinin (HA) and neuraminidase (NA) sequences were acquired from two sources: the Influenza Research Database (IRD) \cite{squires2012influenza} and the Global Initiative on Sharing Avian Influenza Data (GISAID) \cite{shu2017gisaid}.  The initial data collection process yielded 381,369 HA sequences and 338,631 NA sequences (completed on 13 December 2022). ). To maintain the uniqueness of each strain, redundant and multi-label sequences were filtered, resulting in a unique HA and NA sequence pair for each strain in the final data set. To prevent duplicates, the integration process involved removing sequences from GISAID if they were already present in IRD. Additionally, strains belonging to the H0N0 subtype, which have an uncleaved HA0 protein that is not infectious, were also removed from the data set. The process of data curation also involved eliminating sequences with erroneous or ambiguous metadata labels. For example, A/American Pelican/Kansas/W22-200/2022 (isolated ID: EPI\_ISL\_14937098) was inaccurately labelled as ‘host’. Subsequently, the final outcome comprised 46,172 unique pairs of complete and partial HA and NA sequences

The criterion for defining the completeness of A sequence was considered complete if its length was equivalent to that of the actual genomic sequence \cite{shu2017gisaid} or the complete coding region defined by The National Center for Biotechnology Information (NCBI) \cite{squires2012influenza}. The completeness annotation cannot be explicitly obtained from the metadata of the strain. Therefore, incomplete sequences were obtained by filtering the complete sequences from the full influenza database, which comprises both complete and incomplete sequences (\(all \ sequences = complete \ sequences \cup incomplete \ sequences\)). 

The pre-trained model was trained using a training data set comprising sequences of strains isolated before 2020. Conversely, the sequences of strains isolated from 2020 to 2022 were used solely to evaluate the performance of the models during testing; the testing data set also included incomplete sequences. The characteristics of the data sets used in this study are presented in Table~\ref{tab_data}.

\begin{table}\centering
\caption{Summary statistics of data sets.}
\label{tab_data}
\ra{1.3}
\begin{tabular}{@{}cccc@{}}
\toprule
Data Set  \scriptsize(\textit{alias})        & \# Total Pairs & \# Seqs from IRD & \# Seqs from  GISAID \\ \midrule
\textless \ 2020  \scriptsize (\textit{pre-20}) & 33,159  & 41,940   & 24,378     \\
2020 - 2022  \scriptsize(\textit{post-20})    &  4,488   & 3,232  & 5,744       \\
Incomplete \scriptsize (\textit{incomplete})    &  8,525   & 11,111   & 5,939        \\ \bottomrule
\end{tabular}
\end{table}

\subsubsection{Label Reassignment}
\begin{figure*}[ht]\centering
\includegraphics[width= \linewidth]{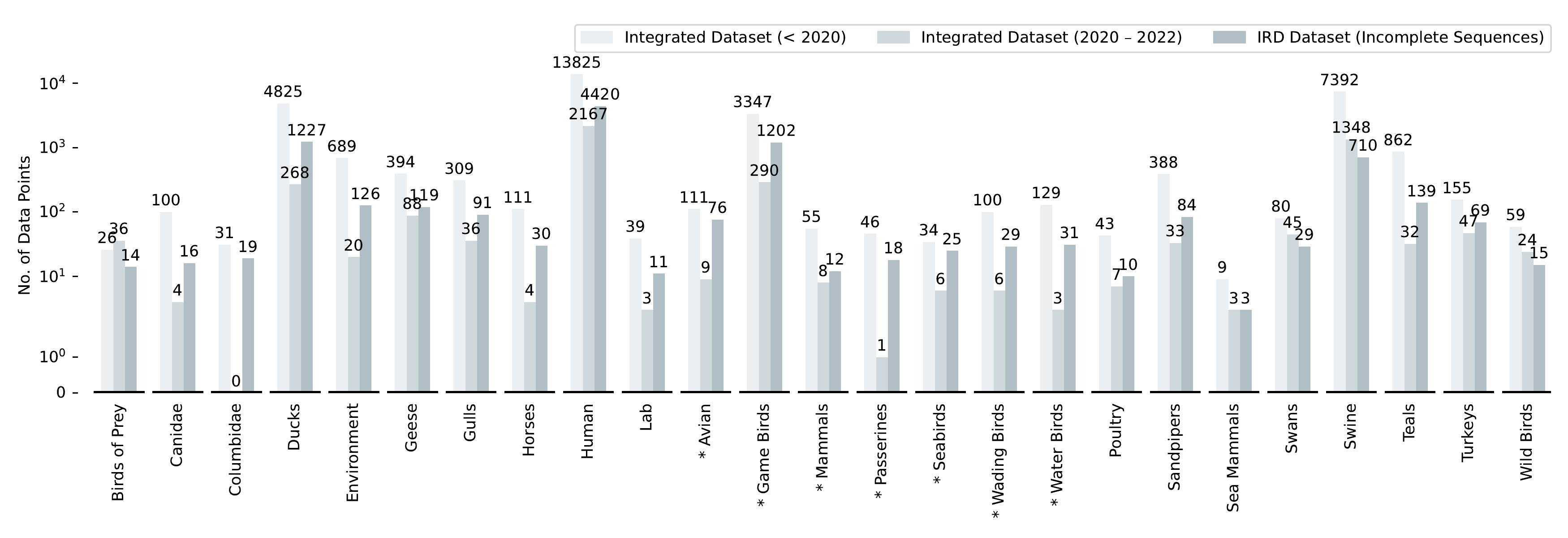}
\caption{Data distribution (hosts)}
\label{fig_data_host}
\end{figure*}

\begin{figure*}[ht]\centering
\includegraphics[width= \linewidth]{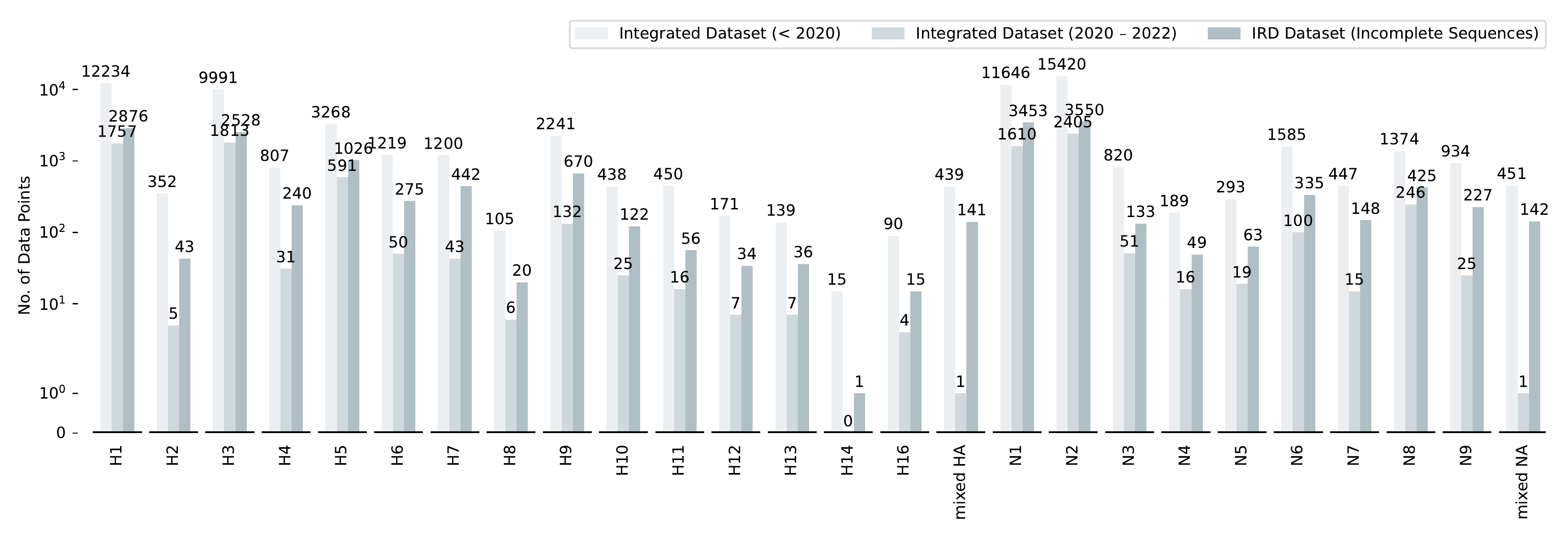}
\caption{Data distribution (subtypes)}
\label{fig_data_hn}
\end{figure*}

While the GISAID and IRD databases recorded $>$ 300 hosts, only 30\% were consistent across both databases. This issue could be attributed to the blended use of animals’ common and scientific names. We regrouped the viral hosts into 25 categories based primarily on the biological family classification of the animals; the distribution of reassigned hosts is presented in  Fig.~\ref{fig_data_host}. We also moved a few subtypes in the data set into other subtypes (i.e. H15, H17, H18, N10, and N11), as shown in Fig.~\ref{fig_data_hn}.

\subsection{Protein Sequence Representation}
Neural networks are mathematical operators that operate on inputs and generate numerical outputs. However, the raw input sequences must be represented as numerical vectors before the neural network can process them. One popular method of vectorising sequences is one-hot encoding. In natural language processing (NLP), the length of the one-hot vector for each word is determined by the size of the vocabulary, which comprises all unique words or tokens in the data. When representing amino acids, the length of the one-hot vector for each amino acid depends on the number of unique amino acids. This results in a sparse matrix for large vocabularies, which is computationally inefficient. An alternative and more powerful approach are to represent each word as a dense vector through word embedding. Word embedding learns the representation of a word by considering its context, allowing similar words to have similar representations. It has been used successfully in the extraction of features from biological sequences \cite{asgari2015continuous}.

The word embedding process can be incorporated into a deep learning model without relying on manually-crafted feature extraction techniques. A protein’s amino acid sequence is usually written as a string of letters but can also be represented as a set of tripeptides, also known as 3-grams. In NLP, \textit{N}-grams refer to \textit{N} consecutive words in a text, and similarly, \textit{N}-grams of a protein sequence refer to \textit{N} consecutive amino acids. For example, the 3-grams of the sequence ”AAADADTICIG” would be ‘AAA’, ‘AAD’, ‘ADA’, ‘DAD’, ‘ADT’, ‘DTI’, ‘TIC’, ‘ICI’, and ‘CIG’. N was set to 3 based on previous research findings \cite{xu2022dive, xu2022predicting}.

\section{Neural Network Architectures}
In this study, we propose a multi-channel neural network (MC-NN) architecture that incorporates two inputs, namely HA trigrams and NA trigrams, and produces three outputs, specifically host, HA subtypes, and NA subtypes. The neural network models utilized in this research encompass bidirectional gated recurrent unit (BiGRU), convolutional neural network (CNN), and transformer.

\subsection{Bidirectional Gated Recurrent Unit}
The Bidirectional Gated Recurrent Unit (BiGRU) is a model designed to handle sequential data by considering both past and future information at each time step. This model is composed of two separate Gated Recurrent Unit (GRU) layers, one for processing the input sequence in the forward direction and the other for processing the input sequence in the backward direction. The outputs of these two layers are then concatenated and utilised for prediction purposes.

GRUs, similar to Long Short-Term Memory (LSTM) units, possess a reset gate and an update gate \cite{cho2014properties}. The reset gate determines the amount of previous information that needs to be forgotten, while the update gate decides the proportion of information to discard and the proportion of new information to incorporate. Due to fewer tensor operations, GRUs are faster in terms of training speed when compared to LSTMs.

The utilisation of a BiGRU provides the advantage of considering both the past and future context at each time step, thereby leading to more informed predictions. This is particularly useful in sequential data processing where context plays a crucial role in prediction accuracy.

\subsection{Transformer}
The Transformer neural network architecture has had a significant impact in the field of NLP \cite{vaswani2017attention}.  It was initially designed to facilitate machine translation, however, the scope of its application can be broadened to encompass other areas such as addressing protein folding dilemmas \cite{grechishnikova2021transformer}. The Transformer architecture serves as the cornerstone for the advancement of contemporary natural language processing models, including BERT \cite{devlin2018bert}, T5 \cite{raffel2019exploring}, and GPT-3 \cite{brown2020language}. One of the most significant benefits that a Transformer possesses over conventional Recurrent Neural Networks (RNNs) is its capability to process data in a parallel manner. This attribute allows for the utilisation of Graphics Processing Units (GPUs) to optimise the speed of processing and effectively handle extensive text sequences.

The Transformer neural network presents a breakthrough in the field of deep learning through its incorporation of positional encoding and self-attention mechanism. The positional encoding feature serves as a means of preserving the word order information in the data, thereby enabling the neural network to learn and understand the significance of the order. The attention mechanism, on the other hand, allows the model to effectively translate words from the source text to the target text by determining their relative importance. The self-attention mechanism, as implied by its name, allows the neural network to focus on its own internal operations and processes. Through this mechanism, the neural network can comprehend the contextual meaning of words by analysing their relationships and interactions with surrounding words. Furthermore, the self-attention mechanism enables the neural network to not only differentiate between words but also reduce computational requirements, thus improving its efficiency.

\subsection{Convolutional Neural Network}
A Convolutional Neural Network (CNN) was designed to work with image and video data. It is an artificial neural network that uses convolutional layers to extract features from raw data. These convolutional layers analyse the spatial relationship between pixels and learn to recognise patterns in the data. The concept behind Convolutional Neural Networks (CNNs) is based on the visual processing mechanism of the human brain, where neurons are selectively activated in response to various features present in an image, such as edges. In CNNs, two primary types of layers are utilised, namely convolution layers and pooling layers. Convolution layers are the core of the CNN architecture, performing convolution operations on the input image and filters. On the other hand, pooling layers perform down-sampling on the image in order to minimise the number of learnable parameters. This study implements one-dimensional convolution layers to process sequence data.

\section{Implementation and Evaluation Methods}
\begin{figure*}[ht]\centering
\includegraphics[width= .95\linewidth]{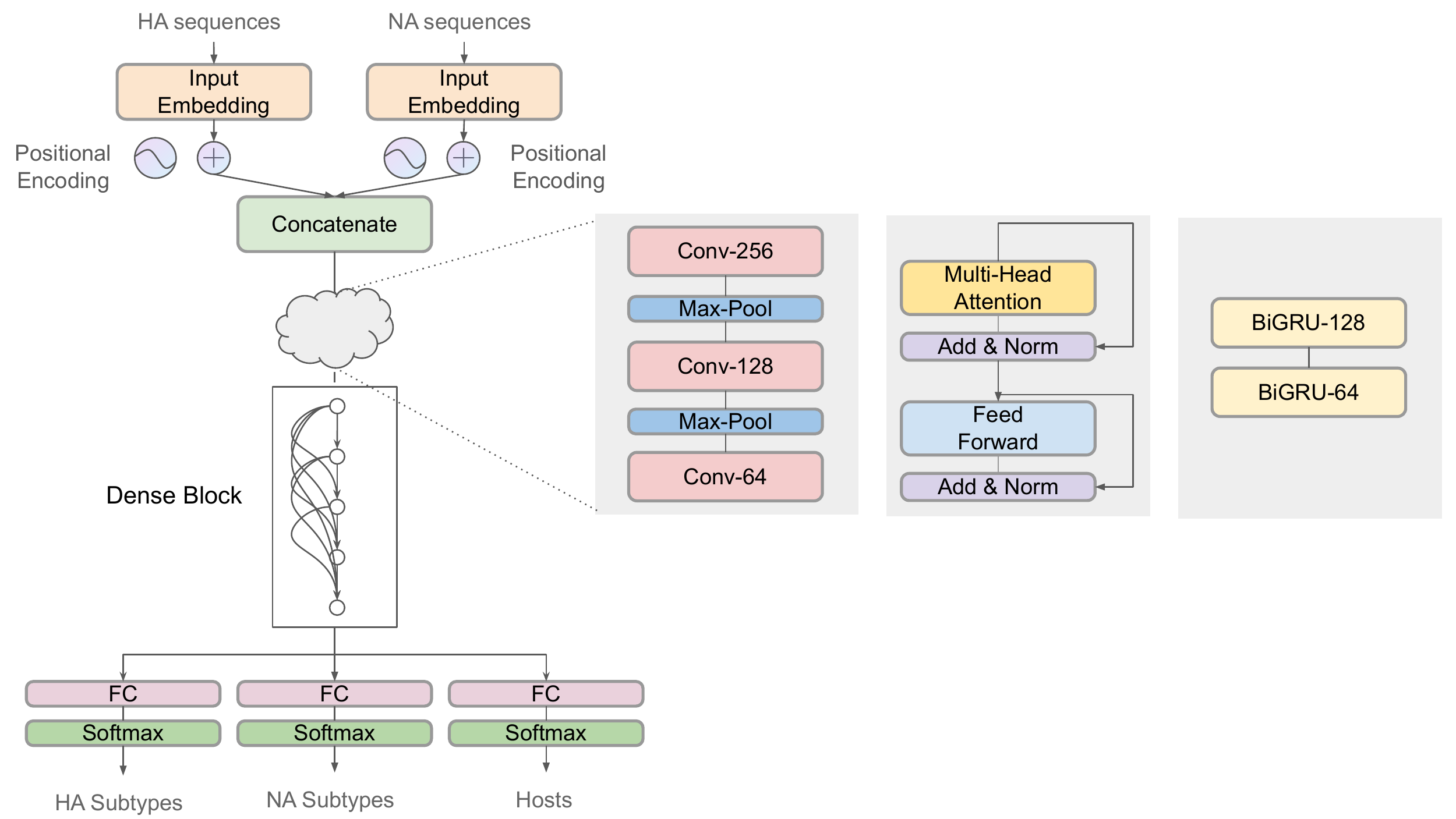}
\caption{The multi-channel neural network architecture: positional encoding  is only employed along with Transformer.}
\label{fig_architecture}
\end{figure*}

All of the models in this study were built using Keras and trained on pre-20 data sets. They were then tested on both post-20 and incomplete data sets. The architecture of the multi-channel neural network used in this study is illustrated in Figure~\ref{fig_architecture}. The Transformer architecture used here is the encoder presented in \cite{vaswani2017attention}.  

In some cases, there is confusion regarding the role of validation and test sets, leading to the tuning of model hyperparameters using the testing set instead of a separate validation set. This increases the risk of data leakage and reduces the credibility of the results. To avoid this issue, nested cross-validation (CV) is used instead of classic K-fold CV. In nested CV, an outer CV is used to estimate the generalisation error of the model and an inner CV is used for model selection and hyperparameter tuning. The outer CV splits the data into a training\textsubscript{outer} set and a testing set, while the inner CV splits the training outer set into a training\textsubscript{inner} set and a validation set. The model is trained only on the training\textsubscript{inner} set, its hyperparameters are tuned based on its performance on the validation set, and its overall performance is evaluated on the testing set. In this study, the outer fold  \(k_{outer}\) was set to 5 and the inner fold \(k_{inner}\)  was set to 4. The hyperparameters settings for the neural network architectures used in this study are presented in Table~\ref{tab_tunning}.,

\begin{table}[]
\caption{Hyperparameter Settings}
\label{tab_tunning} 
\centering
\footnotesize
\ra{1.3}
\begin{tabular*}{\linewidth}{l@{\extracolsep{\fill}}@{}ll@{}}
\toprule
\multicolumn{1}{c}{\textbf{ Models }} & \multicolumn{1}{c}{\textbf{Hyperparameters}} \\
 \midrule
CNN & \begin{tabular}[c]{@{}l@{}}kernel size = 3, 4, 5 \\ embedding size = 50, 100, 150, 200 \\ learning rate = 0.01, 0.005, 0.001, 0.0001  \end{tabular} \\
\hdashline
BiGRU & \begin{tabular}[c]{@{}l@{}} embedding size = 50, 100, 150, 200 \\ learning rate = 0.01, 0.005, 0.001, 0.0001\end{tabular} \\
\hdashline
Transformer & \begin{tabular}[c]{@{}l@{}} embedding size = 32, 64, 128 \\ learning rate = 0.01, 0.005, 0.001, 0.0001 \\ num heads = 1, 2, 3, 4, 5 \end{tabular} \\
\bottomrule
\end{tabular*}
\end{table}
The present study utilises data sets that exhibit a high degree of imbalance, and as such, the application of conventional evaluation metrics such as accuracy and receiver operating characteristic (ROC) curves can lead to misleading results, as demonstrated in prior research \cite{akosa2017predictive, davis2006relationship}.  Precision-recall curve (PRC), on the other hand, has been demonstrated to be more informative when addressing highly imbalanced data sets and has been widely adopted in the research \cite{bunescu2005comparative, bockhorst2005markov, goadrich2004learning, davis2005view}.

The utilisation of linear interpolation to calculate the area under the precision-recall curve (AUPRC) has been shown to be inappropriate \cite{davis2006relationship}. An alternative approach that has been demonstrated to be effective in such cases is the calculation of the average precision (AP) score\cite{su2015relationship}. Furthermore, this study also employs conventional evaluation metrics  F\textsubscript{1} score, with the formulas for these metrics provided below:
\begin{equation}
Precision = \frac{TP}{TP + FP}
\end{equation}
\begin{equation}
Recall = \frac{TP}{TP + FN}
\end{equation}
\begin{equation}
F_1 = 2 \times \frac{Precision \times Recall}{Precision + Recall}
\end{equation}
\begin{equation}
AP =  \sum_{n} (Recall_n - Recall_{n-1} Precision_n)
\end{equation}
where TP, FP, TN, FN stand for true positive, false positive, true negative and false negative. If positive data is predicted as negative, then it counts as FN, and so on for TN, TP and FP.

The evaluation of the overall performance of the models was conducted using the results obtained from the Basic Local Alignment Search Tool (BLAST) as a baseline because BLAST is a commonly employed benchmark in computational biology and bioinformatics.

\section{Results}
\subsection{Overall Performance}
\begin{figure*}[h!]\centering
\includegraphics[width=0.95\linewidth]{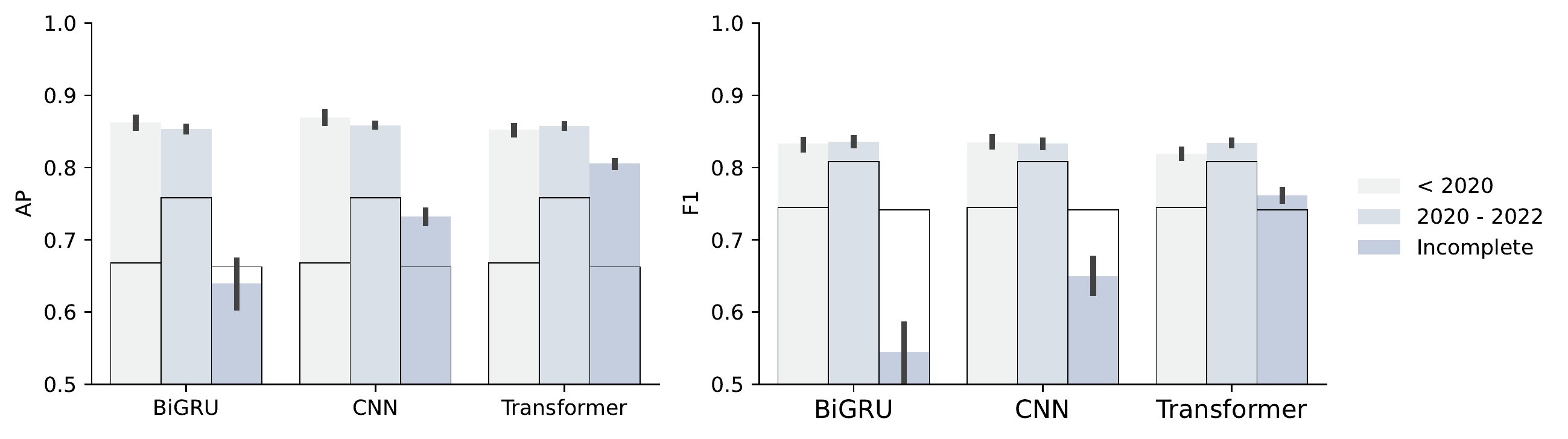}
\caption{Comparison of Overall Performance Between Models (Hosts): the baseline results with BLAST are framed by the black solid line. }
\label{fig_overall_host}
\end{figure*}

\begin{figure*}[h!]\centering
\includegraphics[width=0.95\linewidth]{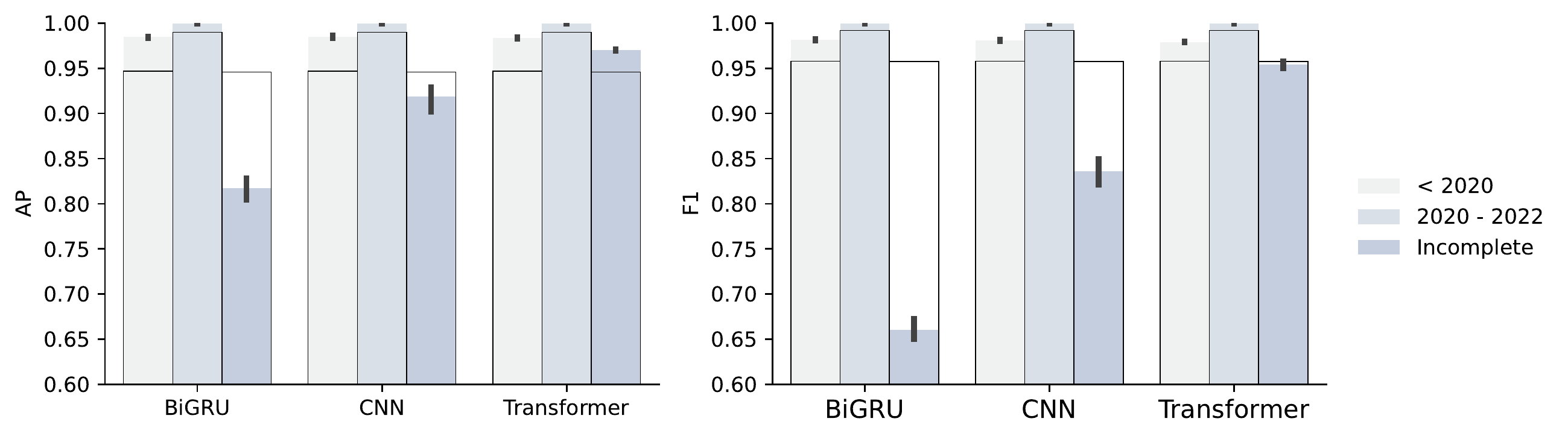}
\caption{Comparison of Overall Performance Between Models (HA subtypes): the baseline results with BLAST are framed by the black solid line.}
\label{fig_overall_ha}
\end{figure*}

\begin{figure*}[]\centering
\includegraphics[width=0.95\linewidth]{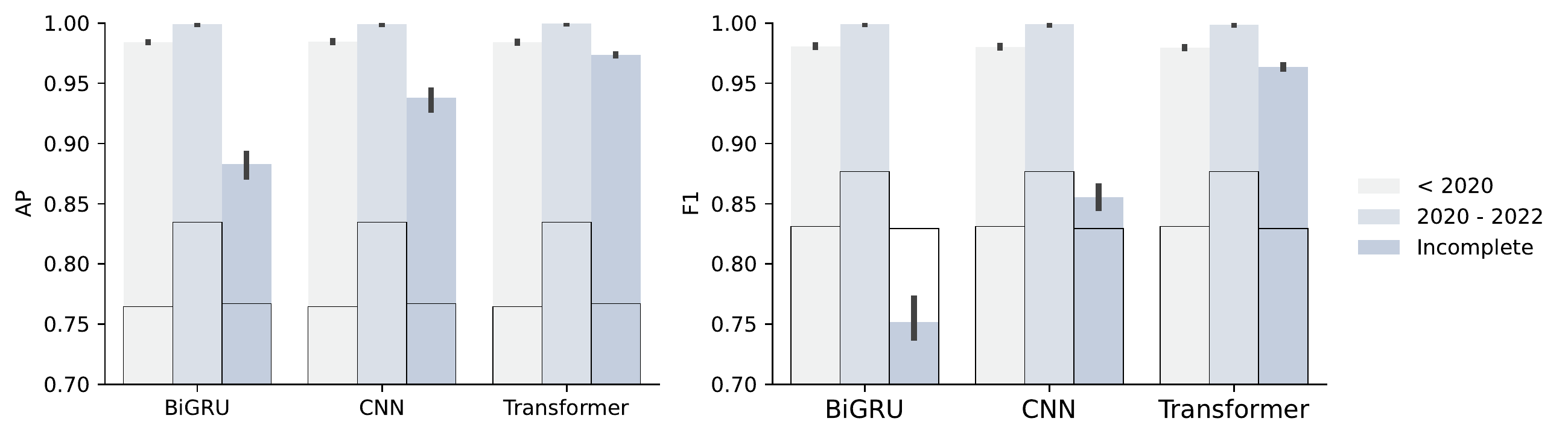}
\caption{Comparison of Overall Performance Between Models (NA Subtypes): the baseline results with BLAST are framed by the black solid line.}
\label{fig_overall_na}
\end{figure*}

The model’s performance on various data sets is shown in Figures~\ref{fig_overall_host} to ~\ref{fig_overall_na}. The metrics used, such as average precision (AP), have been developed for binary classification but can be adapted to multi-class classification using a one-vs-all approach. This approach involves designating one class as positive and all others as negative. AP and F\textsubscript{1} score were used to compare the models to a baseline model, the Basic Local Alignment Search Tool (BLAST), with its default parameters. The BLAST results were obtained through five-fold cross-validation and are indicated by the solid black line in the figures. 

The models were trained solely on the pre-20 data set and tested on the post-20 and incomplete data sets. The pre-20 and post-20 data sets only contained complete sequences, while the incomplete data set included incomplete sequences. All models outperformed the baseline model on the pre-20 data but there was no significant difference in the performance of all models. The results also showed that the host classification task was more challenging than the subtype classification task with all models.

The MC-CNN model exhibited outstanding performance on the pre-20 set across all classification tasks, achieving an average AP of 94.61\% (94.22\%, 94.99\%), and an average F\textsubscript{1} score of 93.20\% (92.86\%, 93.54\%). The MC-CNN model sustained its superior performance on the post-20 set, with an average AP of 95.24\% (95.12\%, 95.36\%), and an average F\textsubscript{1} score of 94.38\% (94.17\%, 94.58\%). The MC-Transformer performed best on the incomplete data set, achieving an average AP of 91.63\% (91.41\%, 91.85\%), and an average F\textsubscript{1} score of 89.29\% (88.80\%, 89.78\%).

\subsection{Performance on Single Sequence Input}
The proposed MC-NN uses two inputs. However, it cannot be guaranteed that the required HA and NA pairs will always be obtainable for every strain. We conducted additional experiments on two data sets, one comprising 23,802 HA protein sequences and the other containing 5,142 NA protein sequences. The results of these experiments are presented in Table~\ref{tab_singleHN}.  The results indicated reduced performance for all models when corresponding H/N sequence pairs were missing. However, the MC-Transformer model outperformed the MC-CNN and MC-BiGRU models on both data sets.

\begin{table}[]
\caption{The overall performance of MC-NN on data sets with single HA or NA sequences.}
\label{tab_singleHN}
\resizebox{\textwidth}{!}{%
\begin{tabular}{@{}ccccc@{}}
\toprule
\multirow{2}{*}{\textbf{Algorithms}} & \multicolumn{2}{c}{\textbf{Single HA}}       & \multicolumn{2}{c}{\textbf{Single NA}}      \\ \cmidrule(l){2-5} 
                                     & \textbf{AP \% (95\% CI) }                & \textbf{F1 \% (95\% CI)}                & \textbf{AP \% (95\% CI)}                & \textbf{F1 \% (95\% CI)}                \\ \midrule
CNN                                  & 76.38 (56.90,  95.86) & 65.04 (55.21, 74.87) & 79.60 (74.84, 84.36) & 62.10 (54.66, 69.55) \\
BiGRU                                & 80.52 (77.26, 83.79)  & 56.26 (46.20, 66.31) & 76.79 (74.96, 78.62) & 59.47 (53.37, 65.57) \\
Transformer                          & \textbf{89.75 (87.96, 91.54)}  & \textbf{76.61(69.79, 83.42)}  & \textbf{83.56 (81.22, 85.91)} & \textbf{70.96 (66.14, 75.78)} \\ \bottomrule
\end{tabular}
}
\end{table}

\section{Conclusion and Discussion}\label{sec12}

The rapid mutation of influenza viruses leads to frequent seasonal outbreaks, although they infrequently result in pandemics. However, these viruses can exacerbate underlying medical conditions, elevating the risk of mortality. In this study, we present a novel approach to predict the viral host at a lower taxonomic level and subtype of the Influenza A virus (IAV) by utilising multi-channel neural networks.

Our approach differs from traditional methods, as it employs a neural network architecture that can learn the embedding of protein trigrams instead of manually encoding protein sequences into numerical vectors. The multi-channel nature of our network eliminates the need for separate models for similar tasks, as it can take multiple inputs and produce multiple outputs. We evaluated the performance of our approach using various algorithms, including CNN, BiGRU, and Transformer, and found that Transformer performed better than the other algorithms. In addition to our previous experiments, we carried out further evaluations to assess the performance of the models in the absence of matching H/N sequence pairs. The results showed that the MC-Transformer model consistently displayed superior performance.

This method could greatly benefit resource-poor regions where laboratory experiments are cost-prohibitive. However, our approach is limited by its reliance on supervised learning algorithms and the need for correctly labelled data, which may result in the poor predictive ability for labels with insufficient data. Further research is needed to address these limitations, including the prediction of cross-species transmissibility and leveraging insufficient data.

\backmatter

\bmhead{Acknowledgments}
The work is supported by the University of Liverpool.

\section*{Declarations}

\textbf{Conflict of interest} The authors have no conflicts of interest to declare. All co-authors have seen and agreed with the contents of the manuscript. We certify that the submission is original work and is not under review at any other publication. \\
\\
\textbf{Ethics approval} This article does not contain any studies with human participants or animals performed by any of the authors.



\bibliographystyle{unsrtnat}
\bibliography{sn-bibliography}

\begin{thebibliography}{10}
\expandafter\ifx\csname url\endcsname\relax
  \def\url#1{\burl{#1}}\fi
\expandafter\ifx\csname urlprefix\endcsname\relax\def\urlprefix{URL }\fi
\providecommand{\bibinfo}[2]{#2}
\providecommand{\eprint}[2][]{\url{#2}}
\providecommand{\doi}[1]{\url{https://doi.org/#1}}
\bibcommenthead

\bibitem{lafond2021global}
\bibinfo{author}{Lafond, K.~E.} \emph{et~al.}
\newblock \bibinfo{title}{Global burden of influenza-associated lower
  respiratory tract infections and hospitalizations among adults: A systematic
  review and meta-analysis}.
\newblock \emph{\bibinfo{journal}{PLoS Medicine}}
  \textbf{\bibinfo{volume}{18}}~(3), \bibinfo{pages}{e1003550}
  (\bibinfo{year}{2021}) .

\bibitem{lau2010viral}
\bibinfo{author}{Lau, L.~L.} \emph{et~al.}
\newblock \bibinfo{title}{Viral shedding and clinical illness in naturally
  acquired influenza virus infections}.
\newblock \emph{\bibinfo{journal}{The Journal of infectious diseases}}
  \textbf{\bibinfo{volume}{201}}~(10), \bibinfo{pages}{1509--1516}
  (\bibinfo{year}{2010}) .

\bibitem{wilde1999effectiveness}
\bibinfo{author}{Wilde, J.~A.} \emph{et~al.}
\newblock \bibinfo{title}{Effectiveness of influenza vaccine in health care
  professionals: a randomized trial}.
\newblock \emph{\bibinfo{journal}{Jama}} \textbf{\bibinfo{volume}{281}}~(10),
  \bibinfo{pages}{908--913} (\bibinfo{year}{1999}) .

\bibitem{watanabe2013renal}
\bibinfo{author}{Watanabe, T.}
\newblock \bibinfo{title}{Renal complications of seasonal and pandemic
  influenza a virus infections}.
\newblock \emph{\bibinfo{journal}{European journal of pediatrics}}
  \textbf{\bibinfo{volume}{172}}~(1), \bibinfo{pages}{15--22}
  (\bibinfo{year}{2013}) .

\bibitem{casas2018aggressive}
\bibinfo{author}{Casas-Aparicio, G.~A.} \emph{et~al.}
\newblock \bibinfo{title}{Aggressive fluid accumulation is associated with
  acute kidney injury and mortality in a cohort of patients with severe
  pneumonia caused by influenza a h1n1 virus}.
\newblock \emph{\bibinfo{journal}{PLoS One}} \textbf{\bibinfo{volume}{13}}~(2),
  \bibinfo{pages}{e0192592} (\bibinfo{year}{2018}) .

\bibitem{public2020influenza}
\bibinfo{author}{England, P.~H.}
\newblock \bibinfo{title}{Influenza: the green book, chapter 19}
  (\bibinfo{year}{2020}) .

\bibitem{shaw2013orthomyxoviridae}
\bibinfo{author}{Shaw, M.} \& \bibinfo{author}{Palese, P.}
\newblock \bibinfo{title}{Orthomyxoviridae, p 1151--1185. fields virology}
  (\bibinfo{year}{2013}).

\bibitem{mills2004transmissibility}
\bibinfo{author}{Mills, C.~E.}, \bibinfo{author}{Robins, J.~M.} \&
  \bibinfo{author}{Lipsitch, M.}
\newblock \bibinfo{title}{Transmissibility of 1918 pandemic influenza}.
\newblock \emph{\bibinfo{journal}{Nature}}
  \textbf{\bibinfo{volume}{432}}~(7019), \bibinfo{pages}{904--906}
  (\bibinfo{year}{2004}) .

\bibitem{asha2019emerging}
\bibinfo{author}{Asha, K.} \& \bibinfo{author}{Kumar, B.}
\newblock \bibinfo{title}{Emerging influenza d virus threat: what we know so
  far!}
\newblock \emph{\bibinfo{journal}{Journal of Clinical Medicine}}
  \textbf{\bibinfo{volume}{8}}~(2), \bibinfo{pages}{192} (\bibinfo{year}{2019})
  .

\bibitem{james2017influenza}
\bibinfo{author}{James, S.~H.} \& \bibinfo{author}{Whitley, R.~J.}
\newblock \bibinfo{title}{ in \textit{Influenza viruses}}
  \bibinfo{pages}{1465--1471} (\bibinfo{publisher}{Elsevier},
  \bibinfo{year}{2017}).

\bibitem{clayville2011influenza}
\bibinfo{author}{Clayville, L.~R.}
\newblock \bibinfo{title}{Influenza update: a review of currently available
  vaccines}.
\newblock \emph{\bibinfo{journal}{Pharmacy and Therapeutics}}
  \textbf{\bibinfo{volume}{36}}~(10), \bibinfo{pages}{659}
  (\bibinfo{year}{2011}) .

\bibitem{vemula2016current}
\bibinfo{author}{Vemula, S.~V.} \emph{et~al.}
\newblock \bibinfo{title}{Current approaches for diagnosis of influenza virus
  infections in humans}.
\newblock \emph{\bibinfo{journal}{Viruses}} \textbf{\bibinfo{volume}{8}}~(4),
  \bibinfo{pages}{96} (\bibinfo{year}{2016}) .

\bibitem{fabijanska2019viral}
\bibinfo{author}{Fabija{\'n}ska, A.} \& \bibinfo{author}{Grabowski, S.}
\newblock \bibinfo{title}{Viral genome deep classifier}.
\newblock \emph{\bibinfo{journal}{IEEE Access}} \textbf{\bibinfo{volume}{7}},
  \bibinfo{pages}{81297--81307} (\bibinfo{year}{2019}) .

\bibitem{scarafoni2019predicting}
\bibinfo{author}{Scarafoni, D.}, \bibinfo{author}{Telfer, B.~A.},
  \bibinfo{author}{Ricke, D.~O.}, \bibinfo{author}{Thornton, J.~R.} \&
  \bibinfo{author}{Comolli, J.}
\newblock \bibinfo{title}{Predicting influenza a tropism with end-to-end
  learning of deep networks}.
\newblock \emph{\bibinfo{journal}{Health security}}
  \textbf{\bibinfo{volume}{17}}~(6), \bibinfo{pages}{468--476}
  (\bibinfo{year}{2019}) .

\bibitem{ahsan2018first}
\bibinfo{author}{Ahsan, R.} \& \bibinfo{author}{Ebrahimi, M.}
\newblock \bibinfo{title}{The first implication of image processing techniques
  on influenza a virus sub-typing based on ha/na protein sequences, using
  convolutional deep neural network}.
\newblock \emph{\bibinfo{journal}{bioRxiv}} \bibinfo{pages}{448159}
  (\bibinfo{year}{2018}) .

\bibitem{xu2017predicting}
\bibinfo{author}{Xu, B.}, \bibinfo{author}{Tan, Z.}, \bibinfo{author}{Li, K.},
  \bibinfo{author}{Jiang, T.} \& \bibinfo{author}{Peng, Y.}
\newblock \bibinfo{title}{Predicting the host of influenza viruses based on the
  word vector}.
\newblock \emph{\bibinfo{journal}{PeerJ}} \textbf{\bibinfo{volume}{5}},
  \bibinfo{pages}{e3579} (\bibinfo{year}{2017}) .

\bibitem{kincaid2018n}
\bibinfo{author}{Kincaid, C.}
\newblock \emph{\bibinfo{title}{N-gram methods for influenza host
  classification}}, \bibinfo{pages}{105--107} (\bibinfo{organization}{The
  Steering Committee of The World Congress in Computer Science, Computer~…},
  \bibinfo{year}{2018}).

\bibitem{attaluri2009integrating}
\bibinfo{author}{Attaluri, P.~K.}, \bibinfo{author}{Chen, Z.},
  \bibinfo{author}{Weerakoon, A.~M.} \& \bibinfo{author}{Lu, G.}
\newblock \emph{\bibinfo{title}{Integrating decision tree and hidden markov
  model (hmm) for subtype prediction of human influenza a virus}},
  \bibinfo{pages}{52--58} (\bibinfo{organization}{Springer},
  \bibinfo{year}{2009}).

\bibitem{eng2014predicting}
\bibinfo{author}{Eng, C.~L.}, \bibinfo{author}{Tong, J.~C.} \&
  \bibinfo{author}{Tan, T.~W.}
\newblock \bibinfo{title}{Predicting host tropism of influenza a virus proteins
  using random forest}.
\newblock \emph{\bibinfo{journal}{BMC medical genomics}}
  \textbf{\bibinfo{volume}{7}}~(3), \bibinfo{pages}{1--11}
  (\bibinfo{year}{2014}) .

\bibitem{kwon2020study}
\bibinfo{author}{Kwon, E.}, \bibinfo{author}{Cho, M.}, \bibinfo{author}{Kim,
  H.} \& \bibinfo{author}{Son, H.~S.}
\newblock \bibinfo{title}{A study on host tropism determinants of influenza
  virus using machine learning}.
\newblock \emph{\bibinfo{journal}{Current Bioinformatics}}
  \textbf{\bibinfo{volume}{15}}~(2), \bibinfo{pages}{121--134}
  (\bibinfo{year}{2020}) .

\bibitem{mock2021vidhop}
\bibinfo{author}{Mock, F.}, \bibinfo{author}{Viehweger, A.},
  \bibinfo{author}{Barth, E.} \& \bibinfo{author}{Marz, M.}
\newblock \bibinfo{title}{Vidhop, viral host prediction with deep learning}.
\newblock \emph{\bibinfo{journal}{Bioinformatics}}
  \textbf{\bibinfo{volume}{37}}~(3), \bibinfo{pages}{318--325}
  (\bibinfo{year}{2021}) .

\bibitem{attaluri2010applying}
\bibinfo{author}{Attaluri, P.~K.}, \bibinfo{author}{Chen, Z.} \&
  \bibinfo{author}{Lu, G.}
\newblock \emph{\bibinfo{title}{Applying neural networks to classify influenza
  virus antigenic types and hosts}}, \bibinfo{pages}{1--6}
  (\bibinfo{organization}{IEEE}, \bibinfo{year}{2010}).

\bibitem{chrysostomou2021classification}
\bibinfo{author}{Chrysostomou, C.}, \bibinfo{author}{Alexandrou, F.},
  \bibinfo{author}{Nicolaou, M.~A.} \& \bibinfo{author}{Seker, H.}
\newblock \bibinfo{title}{Classification of influenza hemagglutinin protein
  sequences using convolutional neural networks}.
\newblock \emph{\bibinfo{journal}{arXiv preprint arXiv:2108.04240}}
  (\bibinfo{year}{2021}) .

\bibitem{sherif2017classification}
\bibinfo{author}{Sherif, F.~F.}, \bibinfo{author}{Zayed, N.} \&
  \bibinfo{author}{Fakhr, M.}
\newblock \bibinfo{title}{Classification of host origin in influenza a virus by
  transferring protein sequences into numerical feature vectors}.
\newblock \emph{\bibinfo{journal}{Int J Biol Biomed Eng}}
  \textbf{\bibinfo{volume}{11}} (\bibinfo{year}{2017}) .

\bibitem{yin2020hopper}
\bibinfo{author}{Yin, R.}, \bibinfo{author}{Zhou, X.}, \bibinfo{author}{Rashid,
  S.} \& \bibinfo{author}{Kwoh, C.~K.}
\newblock \bibinfo{title}{Hopper: an adaptive model for probability estimation
  of influenza reassortment through host prediction}.
\newblock \emph{\bibinfo{journal}{BMC medical genomics}}
  \textbf{\bibinfo{volume}{13}}~(1), \bibinfo{pages}{1--13}
  (\bibinfo{year}{2020}) .

\bibitem{george2019biometric}
\bibinfo{author}{George, A.} \emph{et~al.}
\newblock \bibinfo{title}{Biometric face presentation attack detection with
  multi-channel convolutional neural network}.
\newblock \emph{\bibinfo{journal}{IEEE Transactions on Information Forensics
  and Security}} \textbf{\bibinfo{volume}{15}}, \bibinfo{pages}{42--55}
  (\bibinfo{year}{2019}) .

\bibitem{chen2020multi}
\bibinfo{author}{Chen, Y.} \emph{et~al.}
\newblock \bibinfo{title}{A multi-channel deep neural network for relation
  extraction}.
\newblock \emph{\bibinfo{journal}{IEEE Access}} \textbf{\bibinfo{volume}{8}},
  \bibinfo{pages}{13195--13203} (\bibinfo{year}{2020}) .

\bibitem{cao2019multi}
\bibinfo{author}{Cao, Y.}, \bibinfo{author}{Liu, Z.}, \bibinfo{author}{Li, C.},
  \bibinfo{author}{Li, J.} \& \bibinfo{author}{Chua, T.-S.}
\newblock \bibinfo{title}{Multi-channel graph neural network for entity
  alignment}.
\newblock \emph{\bibinfo{journal}{arXiv preprint arXiv:1908.09898}}
  (\bibinfo{year}{2019}) .

\bibitem{yang2018emotion}
\bibinfo{author}{Yang, Y.}, \bibinfo{author}{Wu, Q.}, \bibinfo{author}{Qiu,
  M.}, \bibinfo{author}{Wang, Y.} \& \bibinfo{author}{Chen, X.}
\newblock \emph{\bibinfo{title}{Emotion recognition from multi-channel eeg
  through parallel convolutional recurrent neural network}},
  \bibinfo{pages}{1--7} (\bibinfo{organization}{IEEE}, \bibinfo{year}{2018}).

\bibitem{kerzel2017haptic}
\bibinfo{author}{Kerzel, M.}, \bibinfo{author}{Ali, M.}, \bibinfo{author}{Ng,
  H.~G.} \& \bibinfo{author}{Wermter, S.}
\newblock \emph{\bibinfo{title}{Haptic material classification with a
  multi-channel neural network}}, \bibinfo{pages}{439--446}
  (\bibinfo{organization}{IEEE}, \bibinfo{year}{2017}).

\bibitem{squires2012influenza}
\bibinfo{author}{Squires, R.~B.} \emph{et~al.}
\newblock \bibinfo{title}{Influenza research database: an integrated
  bioinformatics resource for influenza research and surveillance}.
\newblock \emph{\bibinfo{journal}{Influenza and other respiratory viruses}}
  \textbf{\bibinfo{volume}{6}}~(6), \bibinfo{pages}{404--416}
  (\bibinfo{year}{2012}) .

\bibitem{shu2017gisaid}
\bibinfo{author}{Shu, Y.} \& \bibinfo{author}{McCauley, J.}
\newblock \bibinfo{title}{Gisaid: Global initiative on sharing all influenza
  data--from vision to reality}.
\newblock \emph{\bibinfo{journal}{Eurosurveillance}}
  \textbf{\bibinfo{volume}{22}}~(13), \bibinfo{pages}{30494}
  (\bibinfo{year}{2017}) .

\bibitem{asgari2015continuous}
\bibinfo{author}{Asgari, E.} \& \bibinfo{author}{Mofrad, M.~R.}
\newblock \bibinfo{title}{Continuous distributed representation of biological
  sequences for deep proteomics and genomics}.
\newblock \emph{\bibinfo{journal}{PloS one}}
  \textbf{\bibinfo{volume}{10}}~(11), \bibinfo{pages}{e0141287}
  (\bibinfo{year}{2015}) .

\bibitem{xu2022dive}
\bibinfo{author}{Xu, Y.} \& \bibinfo{author}{Wojtczak, D.}
\newblock \bibinfo{title}{Dive into machine learning algorithms for influenza
  virus host prediction with hemagglutinin sequences}.
\newblock \emph{\bibinfo{journal}{Biosystems}} \textbf{\bibinfo{volume}{220}},
  \bibinfo{pages}{104740} (\bibinfo{year}{2022}) .

\bibitem{xu2022predicting}
\bibinfo{author}{Xu, Y.} \& \bibinfo{author}{Wojtczak, D.}
\newblock \bibinfo{title}{Dive into machine learning algorithms for influenza
  virus host prediction with hemagglutinin sequences}.
\newblock \emph{\bibinfo{journal}{arXiv preprint arXiv:2207.13842}}
  (\bibinfo{year}{2022}) .

\bibitem{cho2014properties}
\bibinfo{author}{Cho, K.}, \bibinfo{author}{van Merrienboer, B.},
  \bibinfo{author}{Bahdanau, D.} \& \bibinfo{author}{Bengio, Y.}
\newblock \bibinfo{title}{On the properties of neural machine translation:
  Encoder-decoder approaches} (\bibinfo{year}{2014}).
\newblock \eprint{1409.1259}.

\bibitem{vaswani2017attention}
\bibinfo{author}{Vaswani, A.} \emph{et~al.}
\newblock \bibinfo{title}{Attention is all you need}.
\newblock \emph{\bibinfo{journal}{Advances in neural information processing
  systems}} \textbf{\bibinfo{volume}{30}} (\bibinfo{year}{2017}) .

\bibitem{grechishnikova2021transformer}
\bibinfo{author}{Grechishnikova, D.}
\newblock \bibinfo{title}{Transformer neural network for protein-specific de
  novo drug generation as a machine translation problem}.
\newblock \emph{\bibinfo{journal}{Scientific reports}}
  \textbf{\bibinfo{volume}{11}}~(1), \bibinfo{pages}{1--13}
  (\bibinfo{year}{2021}) .

\bibitem{devlin2018bert}
\bibinfo{author}{Devlin, J.}, \bibinfo{author}{Chang, M.-W.},
  \bibinfo{author}{Lee, K.} \& \bibinfo{author}{Toutanova, K.}
\newblock \bibinfo{title}{Bert: Pre-training of deep bidirectional transformers
  for language understanding}.
\newblock \emph{\bibinfo{journal}{arXiv preprint arXiv:1810.04805}}
  (\bibinfo{year}{2018}) .

\bibitem{raffel2019exploring}
\bibinfo{author}{Raffel, C.} \emph{et~al.}
\newblock \bibinfo{title}{Exploring the limits of transfer learning with a
  unified text-to-text transformer}.
\newblock \emph{\bibinfo{journal}{arXiv preprint arXiv:1910.10683}}
  (\bibinfo{year}{2019}) .

\bibitem{brown2020language}
\bibinfo{author}{Brown, T.} \emph{et~al.}
\newblock \bibinfo{title}{Language models are few-shot learners}.
\newblock \emph{\bibinfo{journal}{Advances in neural information processing
  systems}} \textbf{\bibinfo{volume}{33}}, \bibinfo{pages}{1877--1901}
  (\bibinfo{year}{2020}) .

\bibitem{akosa2017predictive}
\bibinfo{author}{Akosa, J.}
\newblock \emph{\bibinfo{title}{Predictive accuracy: A misleading performance
  measure for highly imbalanced data}}, Vol.~\bibinfo{volume}{12}
  (\bibinfo{year}{2017}).

\bibitem{davis2006relationship}
\bibinfo{author}{Davis, J.} \& \bibinfo{author}{Goadrich, M.}
\newblock \emph{\bibinfo{title}{The relationship between precision-recall and
  roc curves}}, \bibinfo{pages}{233--240} (\bibinfo{year}{2006}).

\bibitem{bunescu2005comparative}
\bibinfo{author}{Bunescu, R.} \emph{et~al.}
\newblock \bibinfo{title}{Comparative experiments on learning information
  extractors for proteins and their interactions}.
\newblock \emph{\bibinfo{journal}{Artificial intelligence in medicine}}
  \textbf{\bibinfo{volume}{33}}~(2), \bibinfo{pages}{139--155}
  (\bibinfo{year}{2005}) .

\bibitem{bockhorst2005markov}
\bibinfo{author}{Bockhorst, J.} \& \bibinfo{author}{Craven, M.}
\newblock \bibinfo{title}{Markov networks for detecting overlapping elements in
  sequence data}.
\newblock \emph{\bibinfo{journal}{Advances in Neural Information Processing
  Systems}} \textbf{\bibinfo{volume}{17}}, \bibinfo{pages}{193--200}
  (\bibinfo{year}{2005}) .

\bibitem{goadrich2004learning}
\bibinfo{author}{Goadrich, M.}, \bibinfo{author}{Oliphant, L.} \&
  \bibinfo{author}{Shavlik, J.}
\newblock \emph{\bibinfo{title}{Learning ensembles of first-order clauses for
  recall-precision curves: A case study in biomedical information extraction}},
  \bibinfo{pages}{98--115} (\bibinfo{organization}{Springer},
  \bibinfo{year}{2004}).

\bibitem{davis2005view}
\bibinfo{author}{Davis, J.} \emph{et~al.}
\newblock \emph{\bibinfo{title}{View learning for statistical relational
  learning: With an application to mammography.}}, \bibinfo{pages}{677--683}
  (\bibinfo{organization}{Citeseer}, \bibinfo{year}{2005}).

\bibitem{su2015relationship}
\bibinfo{author}{Su, W.}, \bibinfo{author}{Yuan, Y.} \& \bibinfo{author}{Zhu,
  M.}
\newblock \emph{\bibinfo{title}{A relationship between the average precision
  and the area under the roc curve}}, \bibinfo{pages}{349--352}
  (\bibinfo{year}{2015}).

\end{thebibliography}

\end{document}